\documentclass[letterpaper, 10 pt, journal, twoside]{IEEEtran} % Comment this line out if you need a4paper

\usepackage{graphics} % for pdf, bitmapped graphics files
\usepackage{epsfig} % for postscript graphics files
\usepackage{mathptmx} % assumes new font selection scheme installed
\usepackage{amsmath} % assumes amsmath package installed
\usepackage{subfigure}
\usepackage{graphicx}
\usepackage{xcolor}
\usepackage{bm}
\usepackage{multirow}
\usepackage{booktabs} 

\newcommand{\with}[1]{-- \emph{with} {#1}}

\begin{document}

\title{Structure-SLAM: Low-Drift Monocular SLAM in Indoor Environments}

\author{Yanyan Li$^{1*}$, Nikolas Brasch$^{1*}$, Yida Wang$^{1}$, Nassir Navab$^{1,2}$, Federico Tombari$^{1,3}$
\thanks{Manuscript received: February, 24, 2020; Revised June, 17, 2020; Accepted August, 5, 2020.}%Use only for final RAL version
\thanks{This paper was recommended for publication by Editor Sven Behnke upon evaluation of the Associate Editor and Reviewers' comments.} %Use only for final RAL version
\thanks{$^{1}$ Yanyan Li, Nikolas Brasch, Yida Wang, Nassir Navab and Federico Tombari are with Technical University of Munich, Germany; {\tt\footnotesize yanyan.li, nikolas.brasch, yida.wang, nassir.navab @tum.de, tombari@in.tum.de}}
\thanks{$^{2}$ Nassir Navab is with Johns Hopkins University, USA.}
\thanks{$^{3}$ Federico Tombari is with Google, Switzerland.}%
\thanks{$^{*}$ authors are with equal contributions.}%
\thanks{Digital Object Identifier (DOI): see top of this page.}
}

\markboth{IEEE Robotics and Automation Letters. Preprint Version. Accepted August, 2020}
{Li \MakeLowercase{\textit{et al.}}: Structure-SLAM}

\maketitle

%%%%%%%%%%%%%%%%%%%%%%%%%%%%%%%%%%%%%%%%%%%%%%%%%%%%%%%%%%%%%%%%%%%%%%%%%%%%%%%%
\begin{abstract}
 
In this paper a low-drift monocular SLAM method is proposed targeting indoor scenarios, where monocular SLAM often fails due to the lack of textured surfaces. Our approach decouples rotation and translation estimation of the tracking process to reduce the long-term drift in indoor environments.
In order to take full advantage of the available geometric information in the scene, surface normals are predicted by a convolutional neural network from each input RGB image in real-time. First, a drift-free rotation is estimated based on lines and surface normals using spherical mean-shift clustering, leveraging the weak Manhattan World assumption. Then translation is computed from point and line features. Finally, the estimated poses are refined with a map-to-frame optimization strategy. 
The proposed method outperforms the state of the art on common SLAM benchmarks such as ICL-NUIM and TUM RGB-D. 
\end{abstract}

\begin{IEEEkeywords}
SLAM, Visual Learning
\end{IEEEkeywords}

\IEEEpeerreviewmaketitle
%%%%%%%%%%%%%%%%%%%%%%%%%%%%%%%%%%%%%%%%%%%%%%%%%%%%%%%%%%%%%%%%%%%%%%%%%%%%%%%%
\section{Introduction}
% \begin{figure*}
% \subfigure[]{
% \begin{minipage}[b]{0.2\textwidth}
% \includegraphics[scale=0.13]{images/t2.png}
% \includegraphics[scale=0.13]{images/t3.png}
% \end{minipage}}
% \subfigure[]{
% \includegraphics[scale=0.32]{images/t4.png}
% }
% \subfigure[]{
% \includegraphics[scale=0.24]{images/t1.png}
% }
% \caption{Scheme of the proposed system (StructureVO). (a) overview of the proposed system; the left part of (b) shows the original image and the detected features; the right part of (b) shows the vanishing direction vectors and surface normal vectors on the Gausssian sphere.}
% \label{figure1}
% \end{figure*}

% SLAM
%why you need it
Visual Simultaneous Localization and Mapping (V-SLAM) systems are important for autonomous robots and augmented reality, as they are used to estimate poses and reconstruct unknown environments. In numerous SLAM use cases and applications, monocular cameras are the most common sensors in indoor scenarios.
%point and line slam
Indoor environments are often characterized by a lack of textured surfaces, and by irregularly distributed feature points. In particular, low-textured walls, floor and ceiling are difficult to deal with by both state-of-the-art feature-based methods~\cite{mur2015orb} as well as direct methods~\cite{engel14eccv},~\cite{engel2017direct}. For low-textured scenes, SLAM systems combining point and line features have been proposed to target low-textured scenes, e.g. Stereo-PLSLAM~\cite{gomezojeda2017pl-slam:}, PLVO~\cite{lu2015robust}, Mono-PLSLAM~\cite{pumarola2017pl-slam:} and~\cite{proencca2018probabilistic}, extending the working scenarios to low-textured environments with visible structural edges.
% Sequential nature of SLAM leads to drift
Since the map is built from a sequence of input frames, small errors accumulate over time, resulting in drift which affects dense reconstruction by leading to misaligned surfaces and artifacts. 

\begin{figure}
\vspace{4pt}
\centering
\subfigure[Structured scene]{\includegraphics[scale=0.18]{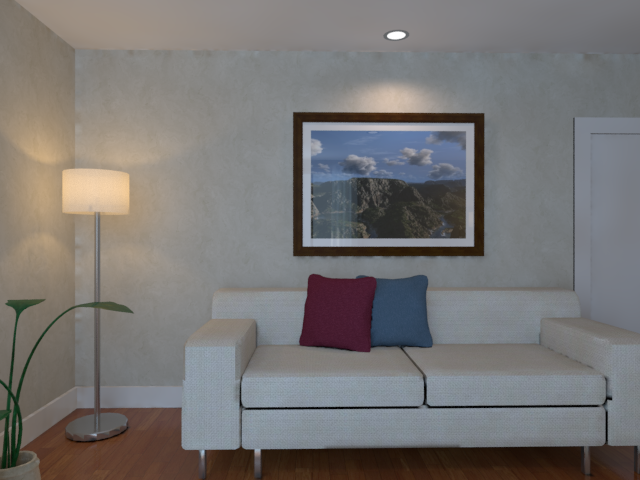}\label{MW_example}}
\subfigure[2D features]
{\includegraphics[scale=0.173]{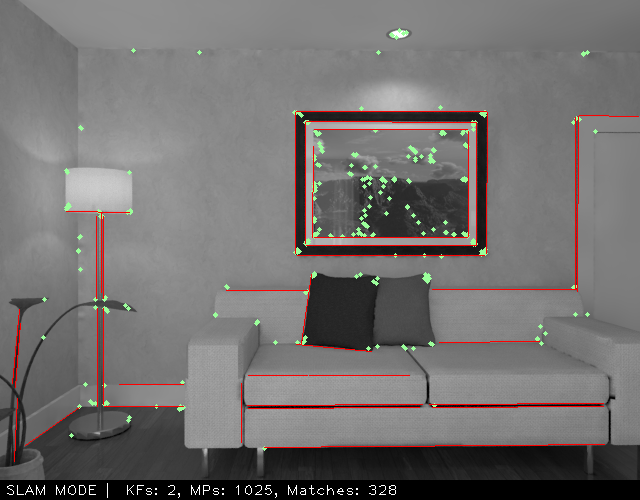}}
\subfigure[Normal prediction]{\includegraphics[scale=0.45]{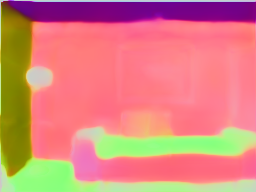}}
\subfigure[plane segmentation from MW]{
\includegraphics[scale=0.173]{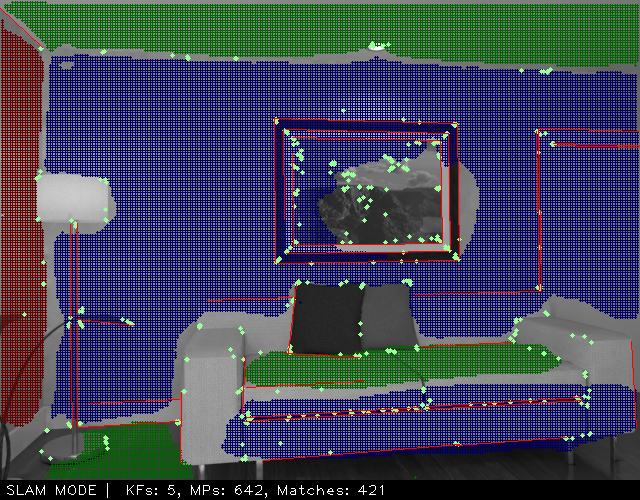}}
\subfigure[Sparse output  map]{\includegraphics[scale=0.32]{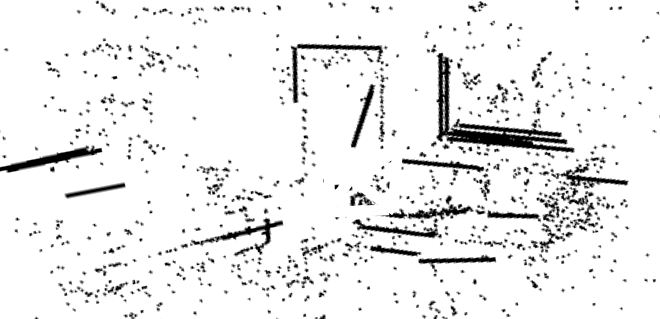}}
%\subfigure[Dense map]{\includegraphics[scale=0.12]{images/t1.png}}
\caption{The proposed approach targets low-textured indoor scenes to carry out low-drift monocular SLAM based on dense normal prediction and leveraging the Manhattan World assumption. %(a) Example of a structured indoor scene (b) Sparse point and line features (c) Dense surface normals of planar areas. (d) plane segmentation. (e) Sparse reconstructed map
}
\label{manhattanworld}
\end{figure}

% Strategies to reduce global drift in SLAM
There are two main strategies to overcome these errors. Loop closure detection~\cite{strasdat2010scale, mur2015orb} combined with pose graph optimization detects previously seen landmarks and optimizes the pose graph based on the new constraints, thus correcting the accumulated drift. Loop closure, however, brings in an extra computational burden and removes the drift only when revisiting the same place. Another strategy consists of assuming an underlying (global) structure in the world frame, then each tracked frame can be directly aligned to this world structure instead of the last frame or keyframes. The most common formulation of a structured scene is the Manhattan World (MW)~\cite{li2018monocular, zhou2015structslam} where the environment shown in Fig.~\ref{MW_example} consists of geometric structures (planes and lines) oriented in one of three orthogonal orientations.
% Example
It is particularly useful in indoor environments where structures such as walls, floor and ceilings often show consistent alignment over multiple rooms, enabling a global alignment.
%Whereas the alignment to global structures can reduce the drift over the whole trajectory, this global structure must exist and be detected robustly.
%But it brings computing burden for the system and cannot be executed until revisit same places. 
% Pros and Cons
% Limitation to a single dominant MW frame and all time visibility
%if this condition fails and another frame is detected the pose can not be recovered later on.
% Mixture of MW frames
%More recent works\cite{TODO} argues that this assumption is too strong and extend the global structure to a combination of Manhattan Worlds, allowing to detect multiple frames and choosing the correct one in a probabilistic way.
% All time visibility
%The methods mentioned above still assume that the global MW frame is visible in all frames, this assumption fails especially in cluttered environments and if the camera moves too close to surface loosing the global context.
% Motivation MW over loop closure
\begin{figure*}
\vspace{5pt}
\centering
\includegraphics[scale=0.4]{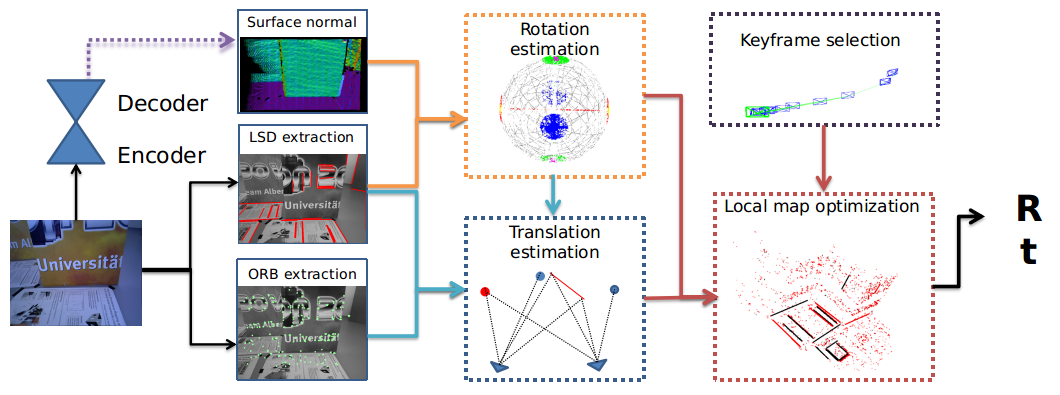}
\caption{Proposed SLAM framework (StructureSLAM). In the front-end, the encoder-decoder network predicts dense surface normals. In parallel, point and line features are extracted from the RGB image. In the back-end, first the scene structure in the form of normals and lines is used to estimate the global rotation of the camera. Then, the remaining 3-DoF for the translation are obtained using point and line features. The initial pose estimate is validated and refined using the local map. Keyframes are selected based on the availability of point and line features.}
\label{figure1}
\end{figure*}

The MW approach is an efficient method to keep the accumulated drift low by providing a drift-free strategy for rotation estimation, as the rotational component is the main source of overall drift~\cite{straub2015real, kim2018low}. %On the other side, existing systems based on the MW assumption will fail if the underlying structure does not correspond to a MW or the global MW structure is not visible in all frames.

% Short comings of state-of-the-art MW approaches
The state of the art of monocular approaches relying on a MW~\cite{li2018monocular},~\cite{zhou2015structslam} are based on parallel and orthogonal lines alone, as it is difficult to extract 3D information, except for vanishing points, from a monocular RGB image, which is a quite strong limitation for most scenarios. Furthermore, indoor environments often consist of large planar regions with few features for pose estimation. RGB-D methods~\cite{kim2018low},~\cite{kim2018linear}, directly measure the structure of the scene in the form of depth maps, this allows them to compute dense surface normals for each pixel.

%On the other hand, the first RGB-D techniques using MW ~\cite{zhou2016divide},~\cite{kim2017visual} are based on the existence of enough orthogonal planes in the environment, which require at least two orthogonal planes to be visible in all frames. This issue is solved by~\cite{kim2018low} and~\cite{kim2018linear} by connecting lines and planes together in the front-end. By relying on frame-to-frame translation estimation using KLT~\cite{bouguet2001pyramidal} to track only good features, drift can still accumulate.

% Learning based normal maps to replace depth/structure sensors
Inspired by recent works based on convolutional neural networks (CNN) and scene geometry prediction approaches from a single view~\cite{yu2019single},~\cite{qi2018geonet}, we propose a monocular SLAM framework which leverages the underlying scene structure to carry out low-drift SLAM even in presence of low-textured environments, in the form of densely predicted normal maps from a CNN, analogously to existing works based on dense RGB-D sensors.

% Contributions
Specifically, we propose the following contributions:
\begin{itemize}
% TODO: What is the difference between these two?
\item A low drift real-time monocular SLAM framework for structured environments, with decoupled rotation and translation
%\item Low drift back-end that combines a decoupled MW rotation and robust translation estimation
\item Dense monocular normal estimation for rotation estimation leveraging the MW assumption
\item A method for translation estimation relying on point and line features
%\item Evaluation on monocular and RGB-D datasets.
%\item Confidence estimation of MW assumption and fallback to non-MW SLAM
\end{itemize}

We evaluate numerically on common SLAM benchmarks such as ICL-NUIM~\cite{handa:etal:ICRA2014} and TUM RGB-D~\cite{sturm2012benchmark} showing that the proposed approach outperforms the state of the art in monocular SLAM. 

\section{Related work}
\subsection{Monocular SLAM}
 PTAM~\cite{klein2009parallel} is a monocular, keyframe-based SLAM system which was the first work to introduce the idea of splitting camera tracking and mapping into parallel threads, and demonstrate to be successful for real time augmented reality applications in small-scale environments. Strasdat et al.~\cite{strasdat2010scale} present a large scale monocular SLAM system in which the front-end bases on optical flow implemented on a GPU, followed by FAST feature matching and motion-only BA, and a back-end based on sliding-window BA. As a complete SLAM pipeline, ORB-SLAM~\cite{mur2015orb} combines feature based tracking, sparse point mapping, descriptor based re-localization and loop closure altogether.
In addition to point features several works propose the use of lines~\cite{gomezojeda2017pl-slam:},~\cite{ lu2015robust} for low-textured environments, we propose to use additional dense structural information in the form of predicted normal maps.

Inspired by the recent success of deep learning based depth prediction, CNN-SLAM~\cite{Tateno2017CNNSLAMRD} incorporates a neural network which estimates depth information within the popular LSD-SLAM~\cite{engel14eccv} framework to create dense scene reconstructions in metric scale, where depth predictions are used to initialize the SLAM system and merged continuously with the semi-dense depth maps optimized by the SLAM system. Instead of estimating depth maps only for key-frames in CNN-SLAM, our approach predicts surface normals from every RGB frame in real-time.
In CodeSLAM~\cite{bloesch2018codeslam}, a neural network learns a compact latent representation for the structure of a scene conditioned on the RGB image, showing that the joint optimization of both structure and pose can improve monocular pose estimation.
By predicting normal maps instead of depth maps we avoid the necessary differentiation operation which could introduce noise. Predicting normal maps also seems to generalize better between datasets as depth does.

\subsection{RGB-D SLAM}
Probabilistic-VO~\cite{proencca2018probabilistic} combines points together with lines and planes for pose estimation while modeling their uncertainties. Due to the combination of 2D-3D point and line correspondences and 3D-3D plane matches, a weighting between re-projection and euclidean errors must be chosen empirically.
%TODO: Maybe mention other papers which sample points on the planes to avoid this issue
%
CPA-SLAM~\cite{ma2016cpa} extended DVO-SLAM with global plane landmarks. Pose estimation and soft assignment of depth measurements to planes are computed in an Expectation-Maximization framework.
KDP-SLAM~\cite{hsiao2017keyframe} combines photometric and geometric loss based on plane segments instead of points for frame-to-frame pose estimation and additionally aligns plane segments with global planes in a Smoothing and Mapping (SAM) framework.
%\textcolor{blue}{In contrast to these works we decouple rotation and translation estimation and use global MW constraints instead of fusing partial planes of arbitrary orientations.}

\subsection{Manhattan World}
% No structure
Straub et al.~\cite{straub2015real} and Zhou et al.~\cite{zhou2016divide} show that the main source of drift in traditional feature-based systems is caused by the rotation estimation. 
%%TODO: Add references or leave it out
%However in many man-made environments, there exists a set of global parallel and orthogonal planes and lines which can be leveraged to determine constraints for the camera pose estimation. Based on these constraints, it is possible to avoid traditional methods that rely on frame-to-frame transformations and align the rotation of every frame to the Manhattan world directly.
% A limitation of the Manhattan assumption is that an absolute orientation estimation is difficult due to the ambiguity between the 3 axes. However, this can be circumvented by choosing the solution closest to the last pose estimate, limiting the motion between frames to less than 45 degrees.

% \begin{figure*}
% \centering
% \includegraphics[scale=0.5]{images/surfaceNormal.pdf}
% \caption{Our proposed architecture for normal from RGB prediction. $3\times3$ convolutions, downsampling and the use of shallow feature representations allow for real-time usage on the CPU.}
% \label{encode-decoder}
% \end{figure*}
Even if the MW assumption is a good constraint for indoor SLAM, it is difficult to enforce it in monocular methods because only limited 3D information can be obtained. Zhou et al.~\cite{zhou2015structslam} applies  J-linkage~\cite{toldo2008robust} to classify parallel line segments into different groups and estimate the dominant direction from the vanishing points. If depth maps are available, surface normals can be computed directly. Joo et al.~\cite{joo2016globally} provide a branch-and-bound framework for Manhattan Frame estimation. MVO~\cite{zhou2016divide} propose a unit sphere mean shift method to find the rotation matrix between the Manhattan World and the camera system. For the translational part, they compute and align density distributions of points in each orthogonal direction, avoiding the costly matching of points.
OPVO~\cite{kim2017visual} use planes to estimate the Manhattan Frame rotation, limiting its application to environments with at least 2 orthogonal planes. LPVO~\cite{kim2018low} adds vanishing points of lines for the rotation estimation. Both use point based methods for translation estimation.
L-SLAM~\cite{kim2018linear} replaces the graph based translation estimation from LPVO with a Kalman filter based SLAM update, using the LPVO translation estimation in the prediction step.
% Difference to existing works
Compared with~\cite{kim2018low, kim2018linear}, we build an initialization module based on points, lines and predicted normals. Further more, a refinement module is added to optimize the pose after the decoupled initialization.

%In this paper, we propose a robust SLAM system for structured environments, which takes advantage of planes, lines and points to estimate rotational and translational motion. We use surface normal vectors and vanishing directions of lines to estimate a drift-free rotation matrix. An initial translation is computed from points and lines matched in the last frame. The initial rotation and translation estimates are then refined using a local map of points and lines from multiple keyframes. In order to obtain accurate vanishing directions of the lines, a novel 3D line fitting method is proposed.

\section{Scene Structure Analysis}
% features we need
The structural information used in the system is analyzed in this section. First, we describe the methods for extraction and triangulation of points and lines; Then, an architecture for surface normal prediction is introduced.
%\begin{figure*}
%\centering
%\includegraphics[scale=0.2]{images/3dfitting.png}
%\caption{Removing outliers from point cloud using RANSAC method.}
%\label{f3}
%\end{figure*}

\subsection{Points and Lines Analysis}
% Points and Lines
Point features, due to their descriptiveness, compactness and robustness to illumination changes, are the most common features used in visual SLAM systems. In our method, ORB features~\cite{rublee2011orb} are adopted which are fast enough to extract and robust enough to get matched. Since it's hard to extract sufficient feature points for robust pose estimation in low-textured environments, we further supplement them with line segments extracted and encoded using the LSD~\cite{von2010lsd} and LBD~\cite{zhang2013efficient} accordingly.

% 3D landmarks
Similar to ORB-SLAM~\cite{mur2015orb}, once the 2D point features $p_n=(u_n,v_n)$ and line segments $l_m=(p_{m,s},p_{m,e})$ are extracted in the new keyframe $F_i$, new features are triangulated to 3D points $P_n$ and lines $L_m$ with correspondences located on other connected keyframes.
%\yida{Unlike what} points \yida{behave, frame $F_i$ where $l_m$ was detected for the first time, would be selected as an} anchor for $l_m$. If its correspondence \yida{gets} detected again, we will triangulate the endpoints of the $l_m$ into the 3D space.

% Initialization
% TODO: Maybe move to initialization section?
Due to the factorization of rotation and translation estimation, it is possible to estimate the pose even in cases with pure rotation and no translation or with small parallax, which would not be possible with pure monocular feature based approaches. The rotation can be estimated from the Manhattan World Frame, this means fewer landmarks are needed to obtain the remaining 3 degrees of freedom for the translation.

\subsection{Surface Normal Prediction}

% Approach
We use learned knowledge to reason about the 3D environment, instead of measuring dense depth values directly. Therefore, a 2D convolutional architecture(CNN) is trained to segment planar regions and predict pixel-wise surface normals.
The proposed CNN is composed of a ResNet101-FPN~\cite{yu2019single} encoder for feature extraction and a two-branch decoder for planar area segmentation and normal estimation. As the planar and non-planar regions are unbalanced in indoor scenarios, we use the balanced cross entropy loss for training
\begin{equation}
    \mathcal{L}_p = -1(1-w)\sum_{i\in P}\log p_i -w\sum_{i\in P_{neg}}\log (1-p_i)~,
\end{equation}
where $P$ and $P_{neg}$ represent planar and non-planar regions, respectively. $p_i$ represents the probability of the $i$th pixel being located in a planar region. We use $w$ to balance the contributions of planar and non-planar pixels.
Then the loss function for the normal estimation is filtered by the planar mask.
\begin{equation}
    \mathcal{L}_n = -\frac{1}{n}\sum_{i\in P} n_i\cdot n_i^*~,
\end{equation}
where $n_i$ and $n_i^*$ are the predicted normal and ground truth normal for the $i$th pixel.

\section{Initialization}
In this section, we describe the strategy of computing the relative poses between two frames and reconstructing an initial map. In order to be robust to different motions, we decouple pose estimation into rotation and translation which is explained further in the following paragraphs.

\begin{figure}
\vspace{5pt}
    \centering
    \includegraphics[scale=0.33]{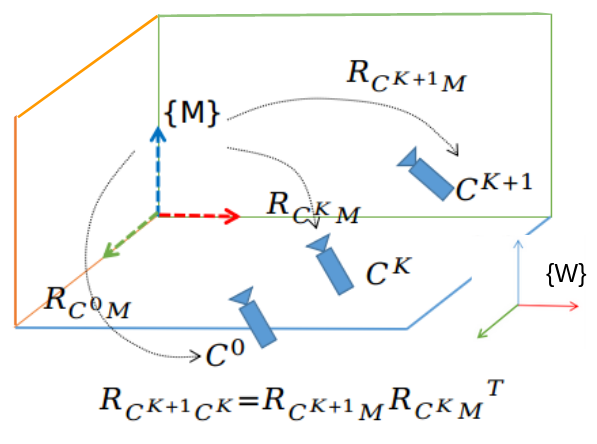}
    \caption{Rotation estimation between multiple frames via the Manhattan world.}
    \label{manhattan_assumption}
\end{figure}

\paragraph{Rotation}
\label{sec:init_rotation}
First, we assume that there is a Manhattan coordinate system ${M}$ shown in Fig.~\ref{manhattan_assumption}, we compute the relative rotation $R_{C_1M}$ from Manhattan coordinate frame ${M}$ to the first frame $C_1$ by clustering the normal map $v_i^s$ of $C_1$ on the unit Gaussian sphere~\cite{zhou2016divide}\cite{kim2018low} centered on the ${M}$. Following ~\cite{zhou2016divide}\cite{kim2018low}, we project the normals onto the tangent plane of each Manhattan world axis $r_n$, where $ n\in[1,2,3]$, for the current estimation. Instead of testing several random matrices, we found that setting $R_{C_0M}$ to identity and running multiple mean-shift iterations is enough to obtain a good estimate. In order to remove noise from normal maps, we only consider the vectors $v_{in}^{s'}$ which are close to the axis $r_n$.

Then, the refined surface normal vectors $v_{in}^{s'}$ are projected to two-dimensional vectors $m_{in}^{'}$ in the $n$th tangential plane. We compute the cluster mean $s_n^{'}$ for the $n$th tangential plane under a Gaussian kernel by
\begin{equation}
s_n^{'}=\frac{\sum_{in}e^{-c\|m_{in}^{'}\|^2} m^{'}_{in}}{\sum_{in}e^{-c\|m_{in}^{'}\|^2}}
\end{equation}
where $c$ is a hyper parameter that defines the width of the kernel, which is set to $2$ in our experiments. Then, we transform the cluster centers back onto the Gaussian sphere as $s_n$, which are used to update the angle between the camera and the MW axis $\hat{r}_n$ combining with the current rotation $Q_n$, 
\begin{equation}
    \hat{r}_n=Q_ns_n
\end{equation}
here $Q_n=[r_{mod(n,3)},r_{mod(n+1,3)}, r_{mod(n+2,3)}]$ and $mod()$ is a modulus operation.
% In order to obtain $R_{C_2M}$, we feed the sphere centered at $R_{C_1M}$ with the normal map of the second frame $C_2$ because two consecutive frames have large overlaps in general.
The tangent plane and the cluster centers are iteratively computed until the rotation estimate is converged. Then we obtain $R_{C_1M}=[\hat{r}_1, \hat{r}_2,\hat{r}_3]^T$.

\paragraph{Translation}
As for the translation estimation, 2D correspondences of points $[p_i^1, p_i^2]$ between two frames and their relative rotation $R_{C_1C_2}$ are used   
\begin{equation}
X_i^2 =\left[ \begin{array}{c}
x_i^2 \\
y_i^2 \\
z_i^2
\end{array} 
\right ]=
\left[ \begin{array}{c}
r_1 \\
r_2 \\
r_3
\end{array} 
\right ]X_i^1 + \left[ \begin{array}{c}
t_1 \\
t_2 \\
t_3
\end{array} 
\right ]
\end{equation}
where $X_i^j$ represents a $3$D point in the $j$th camera. By eliminating the scale $z_i^2$, we obtain
% TODO: Add reference and solve ambiguity between definition of \tilde{x,y}
\begin{equation}
\left[ \begin{array}{c}
\tilde{x}_i^2 \\
\tilde{y}_i^2 \\
1
\end{array} 
\right ] =  \left[ \begin{array}{c}
(r_1 \cdot X_i^1+t_1)/(r_3 \cdot X_i^1+t_3) \\
(r_2 \cdot X_i^1+t_2)/(r_3 \cdot X_i^1+t_3)\\
1
\end{array} 
\right ]
\label{eq:eliminate1}
\end{equation}
where $[ \tilde{x}_i^j \ \tilde{x}_i^j \ 1]^T$ represents the $i$th normalized 3D point in the $j$th camera frame. Since $X_{i}^1$ is also a 3D point, we need to eliminate $z_i^1$ and build
\begin{equation}
   \left[ \begin{array}{c}
-\tilde{y}_i^1t_3+t_2 \\
\tilde{x}_i^1t_3-t_1 \\
-\tilde{x}_i^1t_2+\tilde{y}_i^1t_1
\end{array} 
\right ]^T
 \left[ \begin{array}{c}
r_1 \\
r_2 \\
r_3
\end{array} 
\right ]  \left[ \begin{array}{c}
\tilde{x}_i^1 \\
\tilde{y}_i^1 \\
1
\end{array} 
\right ] = 0
\label{eq:eminate2}
\end{equation}
where $[ \tilde{x}_i^j \ \tilde{x}_i^j \ 1]^T = K^T (u_i^j\ v_i^j\ 1)$ and $K$ is the intrinsic matrix of the camera~\cite{kim2018low}. $(u_i^j\ v_i^j)$ is the $i$th pixel in the $j$th frame. Based on eq. \ref{eq:eliminate1} and eq.\ref{eq:eminate2}, we construct a translation relationship between those 2D correspondences. Then, we solve the system in eq. \ref{eq:eminate2} using SVD to obtain the translation.
 
\section{Tracking}
%explain drift-free rotation
Instead of estimating rotation and translation between two frames, we estimate the rotation between each frame and the underlying Manhattan World. The residual rotation errors are independent of the sequence length and cannot be propagated between frames.
%So our rotation estimation part is a drift-free process compared with traditional feature-based methods.
%translation
Point and line correspondences are used to estimate translation (3 DoFs) by a combination of frame-to-frame and frame-to-map methods.

\subsection{Manhattan Rotation Estimation}\label{MW_estimation}
This section describes the rotation estimation between camera and Manhattan system.

% Mean shift method
% TODO: Add description of planar regions (maybe in the previous section) and how they are extracted (maybe also add vanishing line computation)  ---finished
Given the surface normals and mask of planar regions from the network, we follow the mean-shift clustering approach, as desribed in \ref{sec:init_rotation}, to find the dominant axes on the euclidean sphere and estimate the rotation $R_{C_KM}$. Since normal maps might contain errors due to the network’s inference process, the clustering approach is used to remove outliers first. Furthermore, the initial rotation will be refined in following sections.  
% In the following we will highlight the differences in our approach.
% Initialization
% TODO: Move to initialization? For tracking do you use the last known pose or motion model?
%%--finished
% Rotation estimation
% We follow the same mean-shift method as in the initialization to estimate the rotation.
% TODO: Kind of redundant to the beginning of the paragraph and it is the same as the previous work, so we can also skip it...
%After computing the surface normal vectors $v_k^s$ and vanishing directions $v_k^d$, we project them into the tangent plane of each Manhattan world axis $r_n$ in the euclidean sphere. In order to remove noise, we only consider the vectors $v_{kn}^{s'}$ and $v_{kn}^{d'}$ which are close to the axis $r_n$. In our experiment, the threshold is defined as a conic section with cone $\rho_{th}=10$ degrees. Then, we compute the mean shift in the tangential plane. where $c$ is a design parameter that defines the width of the kernel, which is set to $2$ in our experiments. Then, we transform the cluster centers back onto the Gaussian sphere as $s_n$ to obtain the angle between the camera and the MW axis $r_n$.

\subsection{Translation Estimation}
% Side note: What would happen if we use only points on the Manhattan planes?
After obtaining the rotation matrix, we use the points and line segments to estimate the 3-DoF translational motion, which requires less features than the full 6-DoF estimation. 
We re-project the 3D points from the last frame to the current one and define the error function, based on the re-projection error, as follows,
\begin{equation}
e_{k,j}^p=p_k-\pi( R_{k,j}P_{j}+t_{k,j})
\label{e_p}
\end{equation}
here $\pi ()$ is the projection function. Since the rotation matrix $R_{k,j}$ has already been estimated in the last step, we fix the rotation and only optimize the translation using the right half of the Jacobian matrix for eq. \ref{e_p},

\begin{equation}
\begin{split}
\frac{\partial e_{k,j}^p}{\partial \xi}=\left[{\begin{matrix}
&\frac{xyf_x}{z^2} &-\frac{z^2+x^2}{z^2}f_x & \frac{yf_x}{z} & -\frac{f_x}{z} &0 &\frac{xf_x}{z^2} \\
&\frac{z^2+y^2}{z^2}f_y&-\frac{xyf_y}{z^2}&-\frac{xf_y}{z} &0& -\frac{f_y}{z}&\frac{yf_y}{z^2}
\end{matrix}}\right]
\end{split}
\end{equation}

For the lines we obtain the normalized line function from the 2D endpoints $p_{start}$ and $p_{end}$ as follows,
\begin{equation}
l=\frac{p_{start}\times p_{end}}{\|p_{start}\|\|p_{end}\|}=(a,b,c)
\end{equation}
We formulate the error function based on the point-to-line distance between $l$ and the projected 3D endpoints $P_{start}$ and $P_{end}$ from the matched 3D line in the keyframe. For each endpoint  $P_{x}$, the error function can be noted as,
\begin{equation}
e_{k,j}^l=l \pi( R_{k,j}P_{x}+t_{k,j})
\label{e_l}
\end{equation}

The Jacobian matrix for the line error eq. \ref{e_l} is given by
\begin{equation}
\begin{split}
\frac{\partial e_{k,j}^l}{\partial \xi}=\left[{\begin{matrix}
&-\frac{f_yl_yz^2+f_xl_xxy+f_yl_yy^2}{z^{2}},& \frac{f_xl_xz^2+f_xl_xx^2+f_yl_yxy}{z^2},\\ 
& -\frac{f_xl_xy - f_yl_yx}{z},&\frac{f_xl_x}{z},\\
&\frac{f_yl_y}{z},&  -\frac{f_xl_xx+f_yl_yy}{z^2}
\end{matrix}}\right]
\end{split}
\end{equation}
The combined least squares cost for points and lines can be written as
\begin{equation}
t*=argmin\sum_{j \in(k-2,k-1)}^M ({e_{k,j}^p}^Te_{k,j}^p+{e_{k,P_x}^l}^Te_{k,P_x}^l)
\end{equation}
The system is solved using the Levenberg-Marquardt algorithm.
\subsection{Fallback and Pose Refinement}
% Non Manhattan world/Not visible
The pose estimate is based on the MW assumption. In cases of Non-Manhattan Worlds or where the Manhattan Frame is not visible in the current frame the estimated pose will be incorrect.
%
% Validation of the Initialization
To check whether the pose estimate obtained from the previous steps is correct we project all features from the last $n$ keyframes onto the current frame and compute the re-projection error. By applying a threshold to filter the features we require a minimum number of inliers to accept the pose. 

% RGB-D structure no texture
%If RGB-D information is available

%In this paper, we also build keyframes \cite{murartal2015orb-slam:} that can be used to reconstruct 3D map.

% No good initialization
When not enough inliers are found we fall back to a frame-to-frame tracking method until we estimate a pose that agrees again with the Manhattan World.
% First try to track on last frame
As a fallback we first track the new frame based on the last frame using an efficient re-projection search scheme~\cite{murartal2017orb-slam2:} for points and lines, using the same least squares method as for the translation, this time using the full Jacobian matrix.
% If not successful try last keyframe
In the case we do not get a good solution, measured based on the number of inliers, we try to estimate the pose based on the last keyframe using descriptor matching for the points~\cite{murartal2017orb-slam2:} and re-projection based search for the lines.
% Finally optimize pose based on local map from multiple keyframes
To reduce the drift, in the final step we optimize the pose of the new frame based on a local map constructed from the last $n$ keyframes~\cite{murartal2017orb-slam2:}. Here we do not use the MW assumption anymore, as we found that the initial rotation estimation is enough to reduce the drift and errors in the predicted normal maps can lead to inconsistent pose estimates.

In contrast to other work, based on Manhattan frames for rotation estimation this heuristic allows us to fall back to a purely feature based pose estimation in case the estimate from the MW pose estimation is wrong or not available.

\section{Experiments}

\noindent\textbf{Implementation details } We train the network implemented for normal estimation based on the ScanNet~\cite{dai2017scannet} dataset with a batch size of 32 for 8 epochs. The backbone is pretraind on ImageNet~\cite{deng2009imagenet} for feature extraction and PlaneReconstruction~\cite{yu2019single} for understanding plane regions. We use the Adam optimizer with a learning rate of $10^{-4}$ and a weight decay of $10^{-5}$. Our model is trained in an end-to-end manner and can predict normal maps in real-time. As a baseline we use the original GeoNet~\cite{qi2018geonet} model trained by the authors for 400k iterations on NYU-DepthV2~\cite{Silberman:ECCV12}. Models used in the experiments are not fine-tuned on other datasets.  
% Side note: It is not a very good comparison if the baseline is trained on different data..., but it is too late to change it now ;D
All experiments were carried out with an Intel Core i7-8700 CPU (with @3.20GHz) and a NVIDIA 2080 Ti GPU. We run each sequence 5 times and show median results for the accuracy of the estimated trajectory.
We evaluate our proposed SLAM system on public datasets and compare its performances with other state-of-the-art methods. The evaluation metrics used in the experiments are the absolute trajectory error (ATE) and the relative pose error (RPE)~\cite{sturm2012benchmark}, which measure the absolute and relative pose differences between the estimated and the ground truth motion.
%platform

\noindent\textbf{Evaluation and datasets}
In order to evaluate our method, on the one hand, we compare against several monocular SLAM frameworks, as CNN-SLAM~\cite{Tateno2017CNNSLAMRD} that connects SLAM with predicted depth maps based on keyframes, LSD-SLAM~\cite{engel14eccv} that is popular direct method and ORB-SLAM~\cite{mur2015orb}. We align the trajectories for ORB-SLAM, LSD and the proposed method to the ground truth trajectories using a similarity transformation~\cite{mur2015orb} due to the unknown real scale.
On the other hand, we run our SLAM architecture with different normal maps to evaluate the importance of accurate normals, by switching our normals with the ones from the state-of-the-art, but not real-time capable network, GeoNet~\cite{qi2018geonet} and normal maps computed from the depth maps provided by the dataset using~\cite{feng2014fast}.
% We run our system using normal maps from the  GeoNet~\cite{qi2018geonet} and filtering method~\cite{feng2014fast} that is computed from the provided depth maps.

\begin{itemize}
\item 
ICL-NUIM dataset~\cite{handa:etal:ICRA2014} is a synthetic indoor datasets that provide RGB images, depth maps and ground-truth camera poses. There are  two scenes, named "living room" and "office" which are noted as "lr" and "of" in our experiments. 
\item 
TUM RGB-D dataset~\cite{sturm2012benchmark} was collected using a real RGB-D sensor in real scenes as well as specially designed scenes to challenge current SLAM algorithms, featuring challenging scenes with good structure, but without texture.
% \item TAMU RGB-D~\cite{lu2015robust} contains RGB-D images in larger scale man-made  environments  like corridors and stairs inside a building. This dataset doesn't provide ground-truth but has same start point and end point in each sequence.
\item 
HRBB4 dataset~\cite{lu2014high} which has 12,000 frames of $640\times320$ pixels recorded by a monocular camera in a corridor.  
\end{itemize}
\begin{figure*}[th]
\centering
\vspace{5pt}
%\subfigure[lr-1]{\includegraphics[scale=0.2]{images/lr1.png}}
\includegraphics[scale=0.4]{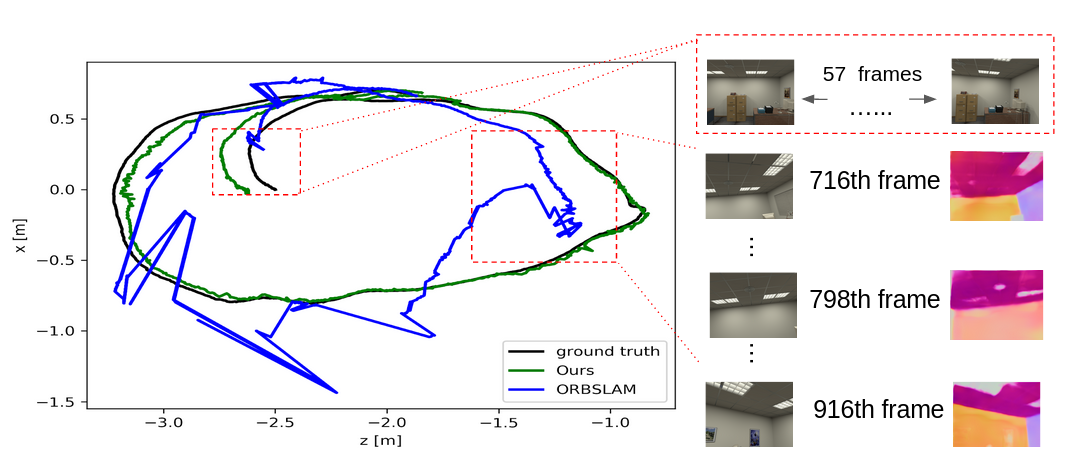}
\caption{Trajectory analysis, comparing the proposed method, ORB-SLAM and the ground-truth on the "of-k3" sequence in the ICL NUIM dataset.}
\label{fig:traj-or3}
\end{figure*}

%%normal prediction
\begin{figure}
\vspace{5pt}
\subfigure[Input]{
\begin{minipage}[b]{0.1\textwidth}
\includegraphics[width=\textwidth]{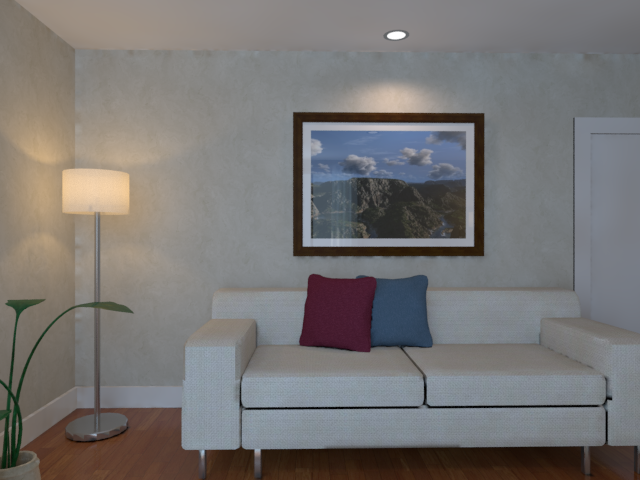} \\
\includegraphics[width=\textwidth]{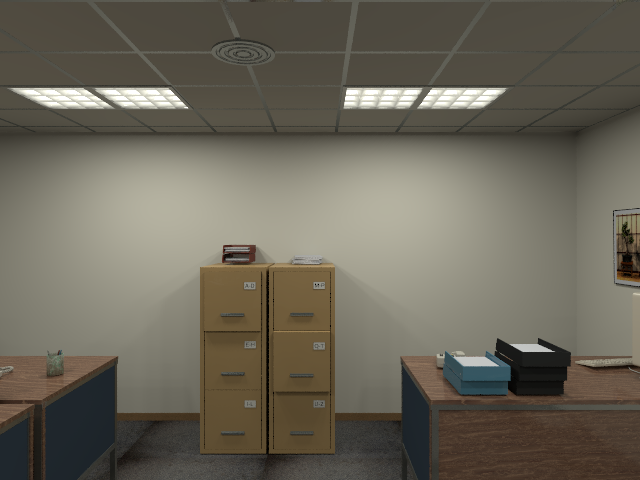}
\includegraphics[width=\textwidth]{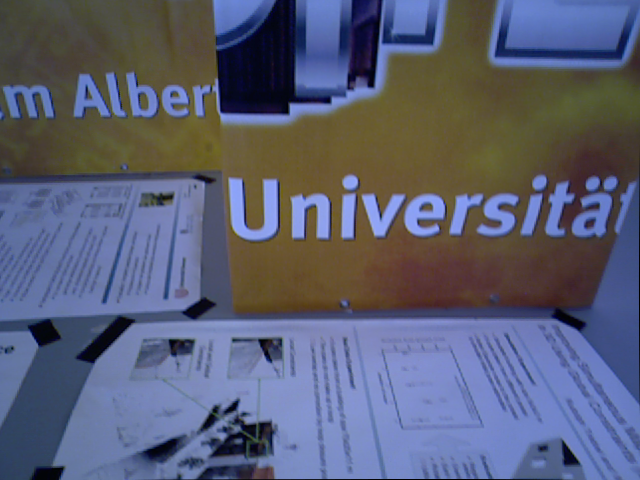}
\includegraphics[width=\textwidth]{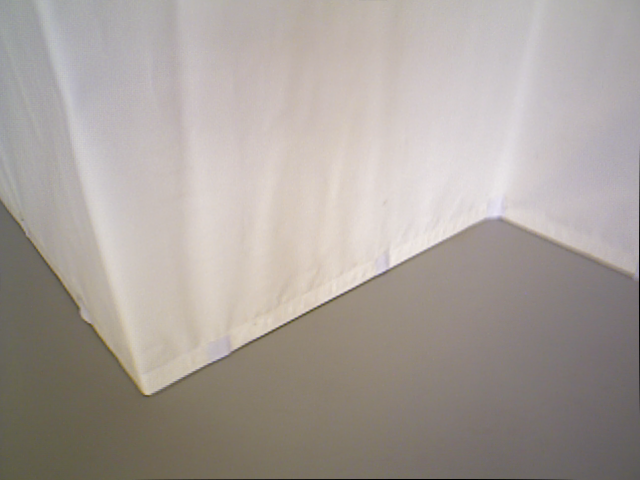}
\end{minipage}
}
\subfigure[Ours]{
\begin{minipage}[b]{0.1\textwidth}
\includegraphics[width=\textwidth]{images/lr-ours.png} \\
\includegraphics[width=\textwidth]{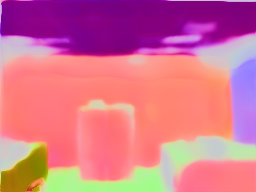}
\includegraphics[width=\textwidth]{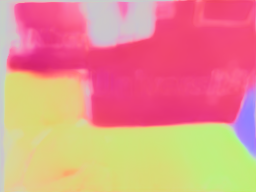}
\includegraphics[width=\textwidth]{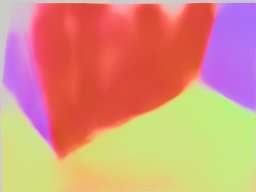}
\end{minipage}
}
\subfigure[GeoNet~\cite{qi2018geonet}]{
\begin{minipage}[b]{0.1\textwidth}
\includegraphics[width=\textwidth]{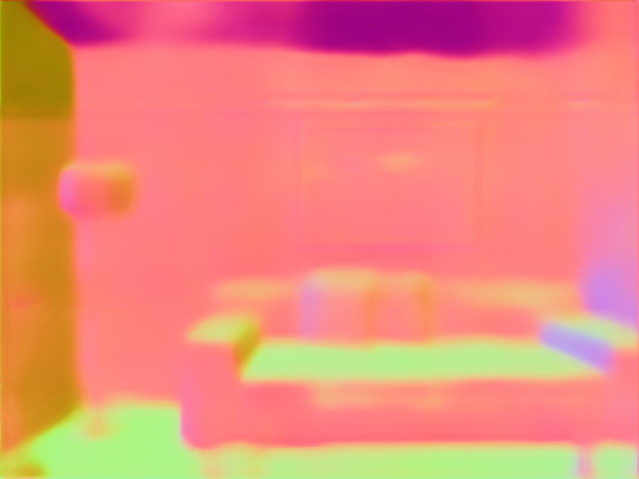} \\
\includegraphics[width=\textwidth]{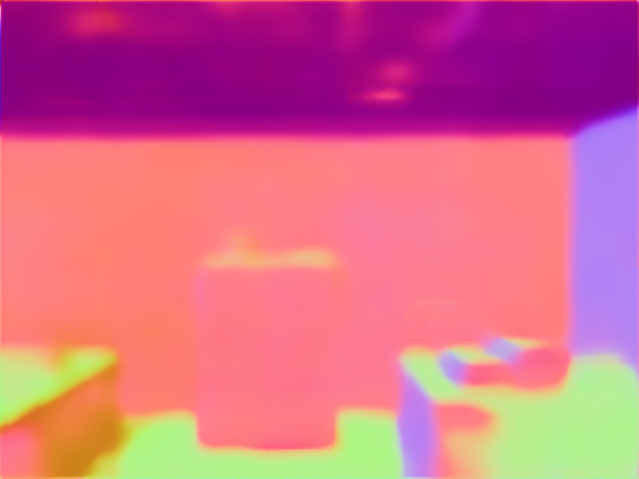}
\includegraphics[width=\textwidth]{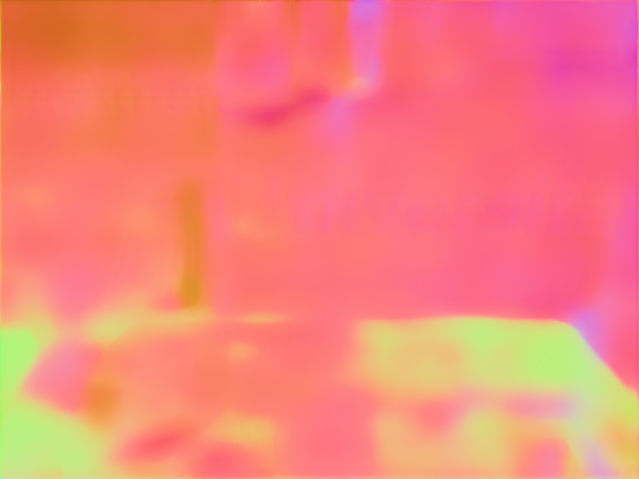}
\includegraphics[width=\textwidth]{images/stnear_geo.png}
\end{minipage}
}
\subfigure[AHC~\cite{feng2014fast}]{
\begin{minipage}[b]{0.1\textwidth}
\includegraphics[width=\textwidth]{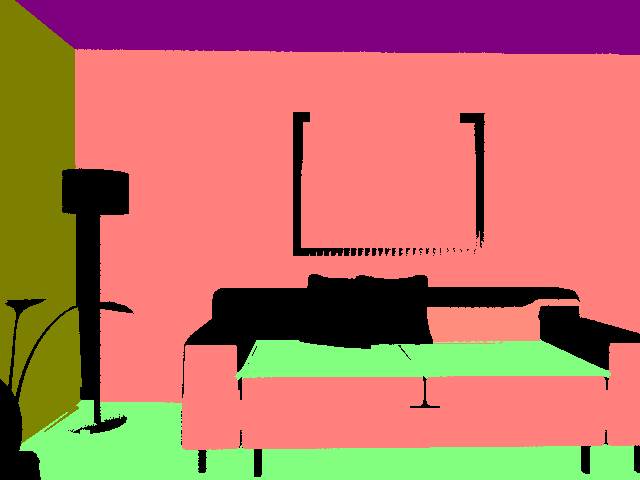} \\
\includegraphics[width=\textwidth]{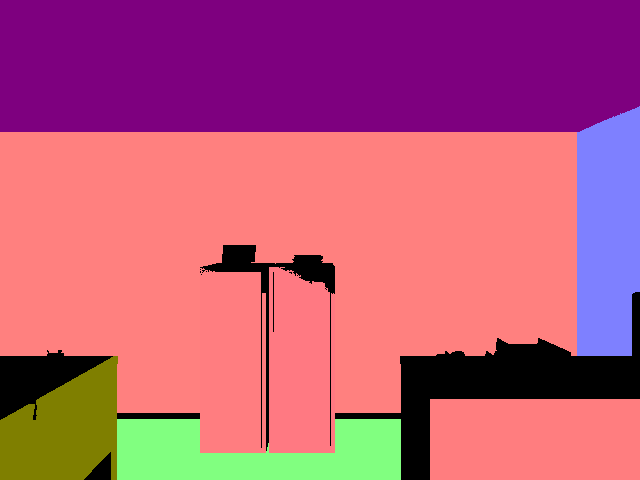}
\includegraphics[width=\textwidth]{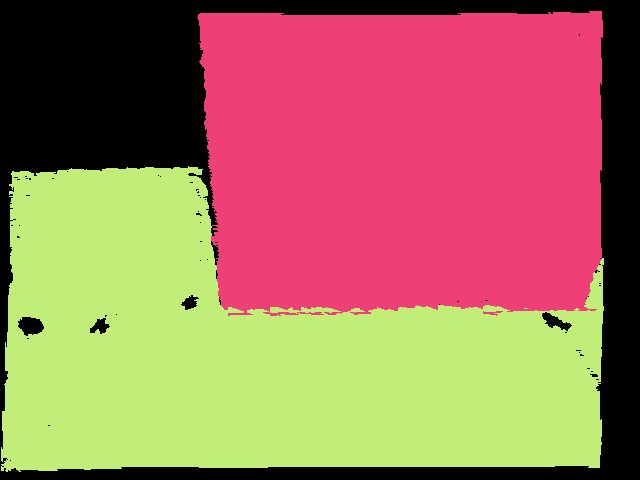}
\includegraphics[width=\textwidth]{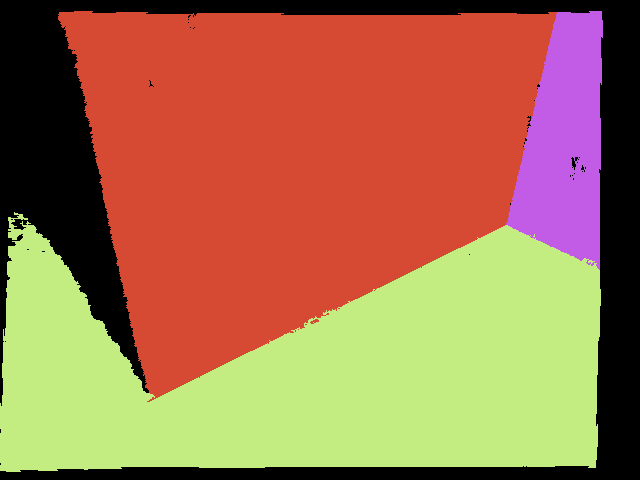}
\end{minipage}
}
\caption{Results of normal prediction model on ICL-NUIM (top) and TUM-RGBD (bottom) scenes for different approaches.}
\label{normalresult}
\end{figure}

\subsection{Normal Prediction}

Fig.\ref{normalresult} presents qualitative results on unseen images of different normal estimation methods. In our method, we mask out the lampshade (first row) and small boxes (second row), as these regions are classified as non-planar. The first two rows, show common examples for indoor environments. Both of them show good results, GeoNet shows smaller inaccuracies. For the last two rows, which are very uncommon scenes, the planar region detection and normal estimation of our model are still generating reasonable results, while the quality of the normal predictions from GeoNet decreased severely. 

The agglomerative hierarchical clustering (AHC) algorithm~\cite{feng2014fast} is an efficient method to detect planes in a depth map. However it difficult to detect planes (like in the third an forth row) where the quality of the depth maps decrease due to a highly slanted surface. In the Table \ref{table:normal}, the performance of the network is evaluated on the ScanNet~\cite{dai2017scannet} dataset generated by \cite{liu2018planenet} against the ground truth. 
\begin{table}[h]
\caption{Performance of the surface normal prediction on the ScanNet~\cite{dai2017scannet} test set.}
\centering
\setlength{\tabcolsep}{3.4mm}{
 \begin{tabular}{|ccc|ccc|}
  %&\multicolumn{6}{c}{\textbf{\normalsize Monocular Camera}} \\ 
	\hline
 \multicolumn{3}{|c|}{\textbf{Error}}  &\multicolumn{3}{c|}{\textbf{Accuracy}}
 \\ \hline
mean& median &rmse & \textless $11.25^{\circ}$&\textless$22.5^{\circ}$ &\textless$30^{\circ}$\\ \hline
$15.2^{\circ}$&$7.8^{\circ}$ &$24.4^{\circ}$ &0.672 &0.797 &0.841 \\
\bottomrule
\hline
\end{tabular}}
\label{table:normal}
\end{table}

\begin{table*}
\caption{Comparison of Translation RMSE (m) for ICL-NUIM~\cite{handa:etal:ICRA2014} and TUM RGB-D~\cite{sturm2012benchmark} sequences using monocular camera. We use \textbf{bold} numbers to mark the best result per sequence. $-w$ means that the proposed framework uses the corresponding surface normals. $\times$ indicates that the algorithm fails due to lost tracking.}
\centering
\resizebox{.95\linewidth}{!}
{
 \begin{tabular}{l|c|c|c|c|c|c|c|c|c|c}
  %&\multicolumn{6}{c}{\textbf{\normalsize Monocular Camera}} \\ 
 \toprule
     Methods & ~~lr-kt1~~ & ~~lr-kt2~~ & ~~lr-kt3~~ & ~~of-kt1~~ & ~~of-kt2~~ & ~~of-kt3~~ & s-t-near & ~~s-t-far~~ & s-not-near & s-not-far \\
 \midrule 
     LSD-SLAM~\cite{engel14eccv} & 0.059 & 0.323 & - & \textbf{0.157} & 0.213 & - & - & 0.214 & - & - \\
     CNN-SLAM~\cite{Tateno2017CNNSLAMRD} & 0.540 & 0.211 & - & 0.790 & 0.172 & - & - & 0.037 & - & - \\
     ORB-SLAM~\cite{mur2015orb} & 0.024 & 0.061 & 0.035 & $\times$ & \textbf{0.031} & 0.326 & 0.016 & 0.015 & $\times$ & $\times$ \\
 \midrule 
 LPVO~\cite{kim2018low} & \textit{0.04} & \textit{0.03} & \textit{0.10} & \textit{0.05} & \textit{0.04} & \textit{0.03} & \textit{0.11} & \textit{0.17} & \textit{0.08} & \textit{0.07} \\
 \midrule
     -$w$ Ours  & \textbf{0.016} & \textbf{0.045} & 0.046 & $\times$ & \textbf{0.031} & 0.065 & \textbf{0.014} & \textbf{0.014} & \textbf{0.065} & \textbf{0.281} \\
     -$w$ GeoNet~\cite{qi2018geonet} & $\times$ & 0.047 & \textbf{0.026} & $\times$ & 0.048 & \textbf{0.043} & 0.107 & \textbf{0.014} & 0.068 & $\times$ \\
     -$w$ AHC~\cite{feng2014fast} & \textit{0.016} & \textit{0.028} & \textit{0.020} & \textit{0.763} &\textit{0.021} & \textit{0.020} & \textit{0.015} & \textit{0.013} & \textit{0.015} & \textit{0.220} \\
 \bottomrule
 \hline
\end{tabular}
}
\label{table:Monocular}
\end{table*}

% \begin{figure}[]
%     \centering
%     \includegraphics[scale=0.55]{images/traj-stfar.png}
%     \caption{Trajectory comparison of the proposed method, ORB-SLAM and the ground-truth for the "st-far"sequence in the TUM-RGBD dataset. }
%     \label{fig:stfar}
% \end{figure}

\begin{figure*}[t]
\centering
\includegraphics[scale=0.37]{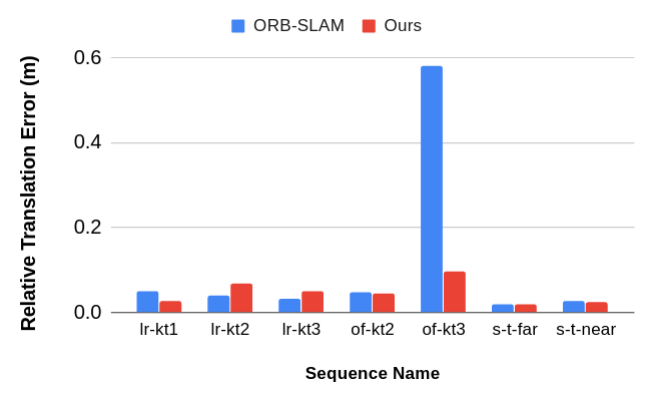}
\includegraphics[scale=0.27]{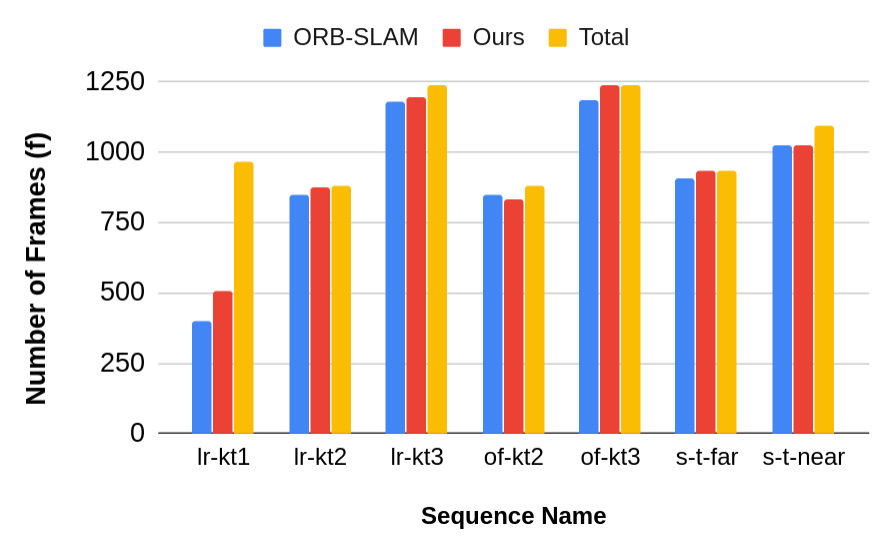}
%\caption{Rotation and translation errors of our method, ORB-SLAM2 and OPVO for the TUM-RGBD "structure-no-texture-far" sequence.}
\caption{Relative translational error comparison between ORB-SLAM and our method on different sequences (left) and a comparison of the average runtime length on each sequence before tracking is lost (right).}
\label{fig:chart}
\end{figure*}

\subsection{Pose Estimation}
%dataset
In order to evaluate our method in different environments, we select structured image sequences from the ICL-NUIM dataset~\cite{handa:etal:ICRA2014} and the TUM RGB-D dataset~\cite{sturm2012benchmark}. 
%table name
Table \ref{table:Monocular} shows the RMSE for all methods on several sequences, 'lr' and 'of' stand for the living room and office room sequences in the ICL-NUIM dataset. 's-t-near' and 's-not-near' are the structure-texture-near and structure-notexture-near sequences in the TUM RGB-D dataset, respectively. 's-t-near' and 's-t-far' are showing the same environment consisting of multiple textured planes, 's-not-near' and 's-not-far' consist of a similar structure, but without texture.

%% analyse monocular results TODO
From the six row to the eight row, different normal maps are given to the same backbone. It is obvious that using AHC-based normal maps (obtained from ground truth depth map) obtain the best results compared to other methods. It also shows the potential of our SLAM architecture, given precise normal maps. Performances from $-w$ Ours (combination of our normal prediction network and the backbone) is more robust than $-w$ GeoNet (combination of GeoNet and the backbone), especially in the 's-not-far' sequence. For those non-textured images, it is difficult for GeoNet to predict accurate normals. In the backbone, conic areas around each axis are used  during the sphere mean-shift method to filter the normal maps, this allows the handling of normal outliers up to a certain point. In cases were the number of outliers is too high, it is difficult to obtain a good rotation from the back-end of the architecture.
Different to the monocular methods, LPVO~\cite{kim2018low} works directly with RGB-D images, which prevents scale drift and allows tracking directly on the depthmap. In comparison our method achieves comparable performance without the use of a depth sensor.

Our method obtains good results and shows robust performance in all five sequences. In the first two sequences, the difference between the point based ORB-SLAM and our method, that connects structure and geometric information, is not significant. However ORB-SLAM is not able to find enough point matches over a sequence of frames and looses tracking in some of the sequences, these are marked with a cross ($\times$). Our method, which additionally uses lines for the translation estimation achieves even better results.

When we compare -$w$ Ours, -$w$ GeoNet with ORB-SLAM in textured sequences, they obtain similar results because those sequences have a sufficient number of features distributed evenly on each frame. However for indoor environments, like Fig.\ref{fig:traj-or3}, it is difficult to obtain enough point features because of large non-textured planar regions. In the 'of-kt3' sequence, there is little change in the first $57$ frames, so ORB-SLAM cannot initialize successfully, because it needs enough points for homography/fundamental model selection. After initialization, it is also challenging for ORB-SLAM to track via the point-based motion model. For our case, the initial rotation matrix is estimated by the mean-shift method instead of estimating the essential or homography matrices. This means we can deal with pure rotational motion. Furthermore, points and line segments are used for 3 DoFs translation only, which is more robust even in large non-textured scene.

%%RPE
\begin{figure}
    \centering
    \includegraphics[scale=0.19]{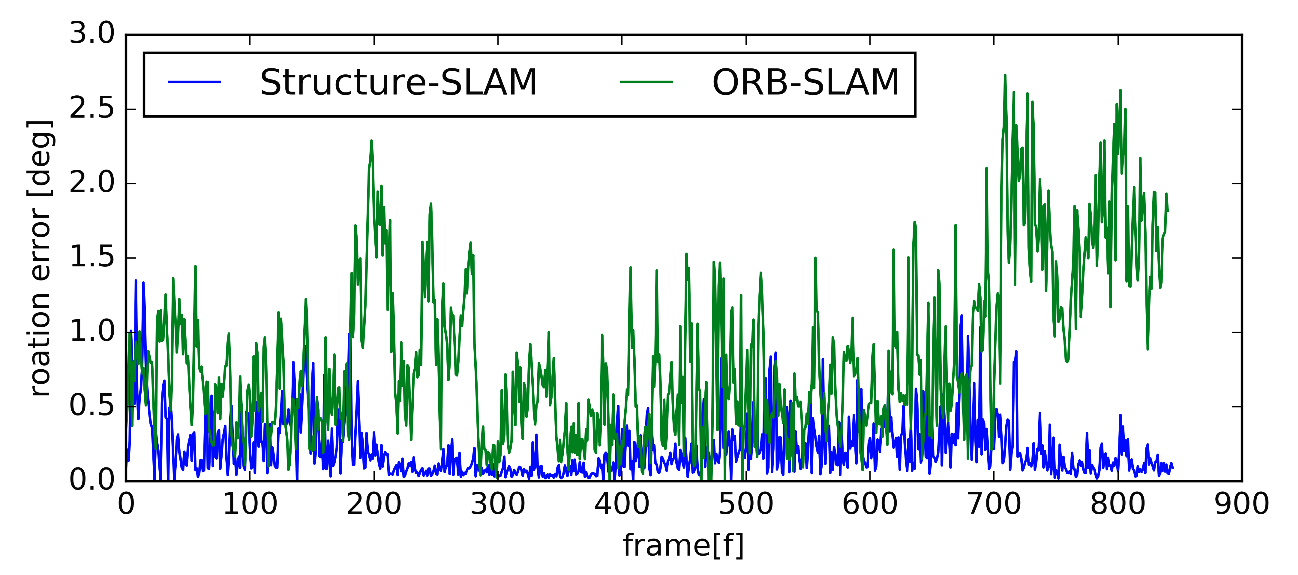}
    \caption{Relative rotation error comparison between ORB-SLAM and our method on sequence lr-k2.}
    \label{fig:rotation}
\end{figure}
%analyse RPE
In order to present the robustness of our method, we compute the RPE for those sequences, which can be processed robustly by ORB-SLAM and our method. For 's-t-far' and 's-t-near' that are textured sequences, ORB-SLAM and the proposed method have similar performences. The relative translation errors for the sequence 'of-kt3' in Fig.~\ref{fig:chart} (left) is significantly larger for ORB-SLAM, which corresponds to the result presented in Fig.~\ref{fig:traj-or3}. As shown in Fig. \ref{fig:rotation}, the proposed method, Structure-SLAM, is more stable in rotation estimation compared with ORB-SLAM.
% TODO: Which figure are you talking about now? I do not see rotational error anywhere anymore

\begin{figure}
    \centering
    \includegraphics[scale=0.35]{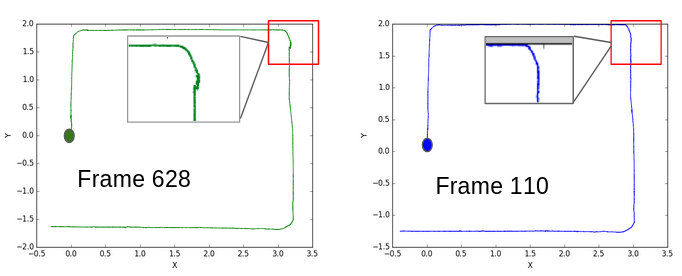}
    \caption{The estimated trajectories of the camera on the HRBB4~\cite{lu2014high} dataset. Left: ORB-SLAM, Right: Structure-SLAM.}
    \label{fig:trj_hrbb4}
\end{figure}

We also compare the number of frames tracked by different methods. Compared with ORB-SLAM, our method retrieves the camera pose more reliable. Especially in 'lr-kt2', 'of-kt3' and 's-t-far', our method initializes fast and tracks all frames in the sequences, as can be seen in sequence 'of-kt3' in Fig. \ref{fig:chart} on the right. Similar results can be found for HRBB4 in Fig. \ref{fig:trj_hrbb4}. Compared with ORB-SLAM which only initializes after the 628th frame, our method is able to initialization much earlier around frame 110. Furthermore, the proposed method shows a more stable behaviour in the upper right corner of the corridor where the environment changes drastically.

% TODO: Which figure now? I do not see one with 5 rows/bars...
% TODO: Also either use ORB-SLAM or ORB-SLAM2, do not mix them that might confuse the reader
% TODO: Add reference to b) part of figure :D

\section{Conclusions}

We have proposed a SLAM system for monocular cameras based on points, lines and surface normals. Using the Manhattan World assumption for rotation estimation and point and line features for windowed translation estimation we achieve state-of-the-art performance. We have shown that normals, learned from a single RGB image, can be used to estimate the rotation between frames leveraging the MW assumption. Compared to other state-of-the-art methods based on global rotation estimation, in our method there exists a fallback level using points and lines to estimate the full pose, in case no Manhattan frame can be found. This enables the tracking over short sequences to later re-localize within the Manhattan world. In the future, global bundle adjustment could be used to correct the frames during these sequences without global frames. Furthermore, we would like to leverage the learned structure information for the translation estimation as well.


\begin{thebibliography}{10}
\providecommand{\url}[1]{#1}
\csname url@samestyle\endcsname
\providecommand{\newblock}{\relax}
\providecommand{\bibinfo}[2]{#2}
\providecommand{\BIBentrySTDinterwordspacing}{\spaceskip=0pt\relax}
\providecommand{\BIBentryALTinterwordstretchfactor}{4}
\providecommand{\BIBentryALTinterwordspacing}{\spaceskip=\fontdimen2\font plus
\BIBentryALTinterwordstretchfactor\fontdimen3\font minus
  \fontdimen4\font\relax}
\providecommand{\BIBforeignlanguage}[2]{{%
\expandafter\ifx\csname l@#1\endcsname\relax
\typeout{** WARNING: IEEEtran.bst: No hyphenation pattern has been}%
\typeout{** loaded for the language `#1'. Using the pattern for}%
\typeout{** the default language instead.}%
\else
\language=\csname l@#1\endcsname
\fi
#2}}
\providecommand{\BIBdecl}{\relax}
\BIBdecl

\bibitem{mur2015orb}
R.~Mur-Artal, J.~M.~M. Montiel, and J.~D. Tardos, ``{ORB-SLAM}: a {V}ersatile
  and {A}ccurate {M}onocular {SLAM} {S}ystem,'' \emph{IEEE Transactions on
  Robotics}, vol.~31, no.~5, pp. 1147--1163, 2015.

\bibitem{engel14eccv}
J.~Engel, T.~Sch\"ops, and D.~Cremers, ``{LSD-SLAM}: {Large-Scale} {D}irect
  {M}onocular {SLAM},'' in \emph{Springer European Conference on Computer
  Vision (ECCV)}, 2014.

\bibitem{engel2017direct}
J.~Engel, V.~Koltun, and D.~Cremers, ``Direct {S}parse {O}dometry,'' \emph{IEEE
  Transactions on Pattern Analysis and Machine Intelligence}, vol.~40, no.~3,
  pp. 611--625, 2017.

\bibitem{gomezojeda2017pl-slam:}
R.~Gomez-Ojeda, F.-A. Moreno, D.~Zu{\~n}iga-No{\"e}l, D.~Scaramuzza, and
  J.~Gonzalez-Jimenez, ``{PL-SLAM}: a {S}tereo {SLAM} {S}ystem {T}hrough the
  {C}ombination of {P}oints and {L}ine {S}egments,'' \emph{IEEE Transactions on
  Robotics}, vol.~35, no.~3, pp. 734--746, 2019.

\bibitem{lu2015robust}
Y.~Lu and D.~Song, ``Robust {RGB-D} {O}dometry {U}sing {P}oint and {L}ine
  {F}eatures,'' in \emph{IEEE International Conference on Computer Vision
  (ICCV)}, 2015.

\bibitem{pumarola2017pl-slam:}
A.~Pumarola, A.~Vakhitov, A.~Agudo, A.~Sanfeliu, and F.~Morenonoguer,
  ``{PL-SLAM}: Real-time {M}onocular {V}isual {SLAM} with {P}oints and
  {L}ines,'' in \emph{IEEE International Conference on Robotics and Automation
  (ICRA)}, 2017.

\bibitem{proencca2018probabilistic}
P.~F. Proen{\c{c}}a and Y.~Gao, ``Probabilistic {RGB-D} {O}dometry {B}ased on
  {P}oints, {L}ines and {P}lanes {U}nder {D}epth {U}ncertainty,''
  \emph{Robotics and Autonomous Systems}, vol. 104, pp. 25--39, 2018.

\bibitem{strasdat2010scale}
H.~Strasdat, J.~Montiel, and A.~J. Davison, ``Scale {D}rift-aware {L}arge
  {S}cale {M}onocular {SLAM},'' \emph{Robotics: Science and Systems VI},
  vol.~2, no.~3, p.~7, 2010.

\bibitem{li2018monocular}
H.~Li, J.~Yao, J.-C. Bazin, X.~Lu, Y.~Xing, and K.~Liu, ``A {M}onocular {SLAM}
  {S}ystem {L}everaging {S}tructural {R}egularity in {M}anhattan {W}orld,'' in
  \emph{IEEE International Conference on Robotics and Automation (ICRA)}, 2018.

\bibitem{zhou2015structslam}
H.~Zhou, D.~Zou, L.~Pei, R.~Ying, P.~Liu, and W.~Yu, ``{StructSLAM}: {V}isual
  {SLAM} with {B}uilding {S}tructure {L}ines,'' \emph{IEEE Transactions on
  Vehicular Technology}, vol.~64, no.~4, pp. 1364--1375, 2015.

\bibitem{straub2015real}
J.~Straub, N.~Bhandari, J.~J. Leonard, and J.~W. Fisher, ``Real-time
  {M}anhattan {W}orld {R}otation {E}stimation in 3{D},'' in \emph{IEEE
  International Conference on Intelligent Robots and Systems (IROS)}, 2015.

\bibitem{kim2018low}
P.~Kim, B.~Coltin, and H.~J. Kim, ``{L}ow-drift {V}isual {O}dometry in
  {S}tructured {E}nvironments by {D}ecoupling {R}otational and {T}ranslational
  {M}otion,'' in \emph{IEEE International Conference on Robotics and Automation
  (ICRA)}, 2018.

\bibitem{kim2018linear}
P.~Kim, B.~Coltin, and H.~Jin~Kim, ``Linear {RGB-D} {SLAM} for {P}lanar
  {E}nvironments,'' in \emph{Springer European Conference on Computer Vision
  (ECCV)}, 2018.

\bibitem{yu2019single}
Z.~Yu, J.~Zheng, D.~Lian, Z.~Zhou, and S.~Gao, ``{Single-Image} {P}iece-wise
  {P}lanar {3D} {R}econstruction via {A}ssociative {E}mbedding,'' in \emph{IEEE
  Conference on Computer Vision and Pattern Recognition (CVPR)}, 2019.

\bibitem{qi2018geonet}
X.~Qi, R.~Liao, Z.~Liu, R.~Urtasun, and J.~Jia, ``{GeoNet}: {G}eometric
  {N}eural {N}etwork for {J}oint {D}epth and {S}urface {N}ormal {E}stimation,''
  in \emph{IEEE Conference on Computer Vision and Pattern Recognition (CVPR)},
  2018.

\bibitem{handa:etal:ICRA2014}
A.~Handa, T.~Whelan, J.~McDonald, and A.~Davison, ``A {B}enchmark for {RGB-D}
  {V}isual {O}dometry, {3D} {R}econstruction and {SLAM},'' in \emph{IEEE
  International Conference on Robotics and Automation (ICRA)}, 2014.

\bibitem{sturm2012benchmark}
J.~Sturm, N.~Engelhard, F.~Endres, W.~Burgard, and D.~Cremers, ``A {B}enchmark
  for the {E}valuation of {RGB-D} {SLAM} {S}ystems,'' in \emph{IEEE
  International Conference on Intelligent Robots and Systems (IROS)}, 2012.

\bibitem{klein2009parallel}
G.~Klein and D.~Murray, ``Parallel {T}racking and {M}apping on a {C}amera
  {P}hone,'' in \emph{IEEE International Symposium on Mixed and Augmented
  Reality}, 2009.

\bibitem{Tateno2017CNNSLAMRD}
K.~Tateno, F.~Tombari, I.~Laina, and N.~Navab, ``{CNN-SLAM}: {R}eal-time
  {D}ense {M}onocular {SLAM} with {L}earned {D}epth {P}rediction,'' in
  \emph{IEEE Conference on Computer Vision and Pattern Recognition (CVPR)},
  2017.

\bibitem{bloesch2018codeslam}
M.~Bloesch, J.~Czarnowski, R.~Clark, S.~Leutenegger, and A.~J. Davison,
  ``{CodeSLAM}--{L}earning a {C}ompact, {O}ptimisable {R}epresentation for
  {D}ense {V}isual {SLAM},'' in \emph{IEEE Conference on Computer Vision and
  Pattern Recognition (CVPR)}, 2018.

\bibitem{ma2016cpa}
L.~Ma, C.~Kerl, J.~St{\"u}ckler, and D.~Cremers, ``{CPA-SLAM}: {C}onsistent
  {P}lane-model {A}lignment for {D}irect {RGB-D} {SLAM},'' in \emph{IEEE
  International Conference on Robotics and Automation (ICRA)}, 2016.

\bibitem{hsiao2017keyframe}
M.~Hsiao, E.~Westman, G.~Zhang, and M.~Kaess, ``Keyframe-based {D}ense {P}lanar
  {SLAM},'' in \emph{IEEE International Conference on Robotics and Automation
  (ICRA)}, 2017.

\bibitem{zhou2016divide}
Y.~Zhou, L.~Kneip, C.~Rodriguez, and H.~Li, ``Divide and {C}onquer: {E}fficient
  {D}ensity-based {T}racking of {3D} {S}ensors in {M}anhattan {W}orlds,'' in
  \emph{Asian Conference on Computer Vision (ACCV)}, 2016.

\bibitem{toldo2008robust}
R.~Toldo and A.~Fusiello, ``Robust {M}ultiple {S}tructures {E}stimation with
  {J-linkage},'' in \emph{Springer European Conference on Computer Vision
  (ECCV)}, 2008.

\bibitem{joo2016globally}
K.~Joo, T.-H. Oh, J.~Kim, and I.~So~Kweon, ``Globally {O}ptimal {M}anhattan
  {F}rame {E}stimation in {R}eal-time,'' in \emph{IEEE Conference on Computer
  Vision and Pattern Recognition (CVPR)}, 2016.

\bibitem{kim2017visual}
P.~Kim, B.~Coltin, and H.~J. Kim, ``Visual {O}dometry with {D}rift-free
  {R}otation {E}stimation {U}sing {I}ndoor {S}cene {R}egularities,'' in
  \emph{British Machine Vision Conference (BMVC)}, 2017.

\bibitem{rublee2011orb}
E.~Rublee, V.~Rabaud, K.~Konolige, and G.~R. Bradski, ``{ORB}: an efficient
  alternative to {SIFT} or {SURF},'' in \emph{IEEE International Conference on
  Computer Vision (ICCV)}, 2011.

\bibitem{von2010lsd}
R.~G. Von~Gioi, J.~Jakubowicz, J.-M. Morel, and G.~Randall, ``{LSD}: A {F}ast
  {L}ine {S}egment {D}etector with a {F}alse {D}etection {C}ontrol,''
  \emph{IEEE Transactions on Pattern Analysis and Machine Intelligence},
  vol.~32, no.~4, pp. 722--732, 2010.

\bibitem{zhang2013efficient}
L.~Zhang and R.~Koch, ``An efficient and robust line segment matching approach
  based on {LBD} descriptor and pairwise geometric consistency,'' \emph{Journal
  of Visual Communication and Image Representation}, vol.~24, no.~7, pp.
  794--805, 2013.

\bibitem{murartal2017orb-slam2:}
R.~Murartal and J.~D. Tardos, ``{ORB-SLAM2}: An {Open-Source} {SLAM} {S}ystem
  for {M}onocular, {S}tereo, and {RGB-D} {C}ameras,'' \emph{IEEE Transactions
  on Robotics}, vol.~33, no.~5, pp. 1255--1262, 2017.

\bibitem{dai2017scannet}
A.~Dai, A.~X. Chang, M.~Savva, M.~Halber, T.~Funkhouser, and M.~Nie{\ss}ner,
  ``{ScanNet}: Richly-annotated {3D} {R}econstructions of {I}ndoor {S}cenes,''
  in \emph{IEEE Conference on Computer Vision and Pattern Recognition (CVPR)},
  2017.

\bibitem{deng2009imagenet}
J.~Deng, W.~Dong, R.~Socher, L.-J. Li, K.~Li, and L.~Fei-Fei, ``{ImageNet}: A
  {L}arge-scale {H}ierarchical {I}mage {D}atabase,'' in \emph{IEEE Conference
  on Computer Vision and Pattern Recognition (CVPR)}, 2009.

\bibitem{Silberman:ECCV12}
P.~K. Nathan~Silberman, Derek~Hoiem and R.~Fergus, ``Indoor {S}egmentation and
  {S}upport {I}nference from {RGBD} {I}mages,'' in \emph{Springer European
  Conference on Computer Vision (ECCV)}, 2012.

\bibitem{feng2014fast}
C.~Feng, Y.~Taguchi, and V.~R. Kamat, ``Fast {P}lane {E}xtraction in
  {O}rganized {P}oint {C}louds {U}sing {A}gglomerative {H}ierarchical
  {C}lustering,'' in \emph{IEEE International Conference on Robotics and
  Automation (ICRA)}, 2014.

\bibitem{lu2014high}
Y.~Lu, D.~Song, and J.~Yi, ``High level landmark-based visual navigation using
  unsupervised geometric constraints in local bundle adjustment,'' in
  \emph{IEEE International Conference on Robotics and Automation (ICRA)}, 2014.

\bibitem{liu2018planenet}
C.~Liu, J.~Yang, D.~Ceylan, E.~Yumer, and Y.~Furukawa, ``{PlaneNet}:
  {P}iece-wise {P}lanar {R}econstruction from a {S}ingle {RGB} {I}mage,'' in
  \emph{EEE Conference on Computer Vision and Pattern Recognition (CVPR)},
  2018.

\end{thebibliography}
\end{document}